\documentclass{article}
\usepackage{spconf,amsmath,amssymb,graphicx,booktabs}
\usepackage[utf8]{inputenc}
\usepackage[T1]{fontenc}
\usepackage{times}
\usepackage[hidelinks]{hyperref}

\title{CORDIAL: Calibrating Ordinal LLM Outputs from Few Labels}
\name{Xiangwei Wang$^{\heartsuit}$, Peng Wang$^{\clubsuit}$, Saman Halgamuge$^{\heartsuit}$}
\address{
$^{\heartsuit}$The University of Melbourne, Australia \quad
$^{\clubsuit}$Shanghai Jiao Tong University, China
}

\newcommand{\ind}{\mathbf{1}}

\newcommand{\softmax}{\operatorname{softmax}}

\begin{document}
\maketitle
\begin{abstract}
A large language model (LLM) can turn a text into a distribution over an ordered scale, but that distribution is a noisy measurement: saturated, compressed or exaggerated, and biased in a consistent direction. We propose CORDIAL, which treats the model’s output as a noisy reading of the true label and corrects it with a channel of five interpretable parameters. The channel is small enough for its posterior to be averaged from a handful of labels, and we prove that the resulting calibration preserves first-order stochastic order. On Amazon reviews and CMU-MOSEI transcripts with four LLMs, CORDIAL has the lowest log loss among nine calibrators in 76 of 80 settings with 5 to 100 labels; with 20 labels and the main 7B reader, it matches the strongest baseline using 28–54 labels. The same posterior lets us learn priors from other tasks and fuse several LLMs. Unrestricted calibrators such as Dirichlet calibration overtake it only as the calibration set grows into the hundreds or thousands.
\end{abstract}

\section{Introduction}
\label{sec:introduction}

A large language model (LLM) can read a review or a spoken sentence and return a distribution over an ordered scale~\cite{kadavath2022language}. Such outputs are useful measurements but not calibrated statistics: a model may understand every sentence and still rate one star too generously~\cite{zhao2021calibrate,zheng2023judging,zhang2023sentiment}, spread its mass too widely or too narrowly~\cite{zhang2026auditing}, or deserve less trust than its confidence suggests, since post-training degrades calibration~\cite{openai2023gpt4,tian2023just}. Treating the output as a noisy measurement of the true label turns calibration into channel estimation from very few labels, which an LLM is deployed to avoid collecting.

The usual answer is a scalar, a temperature or a reliability weight toward a population prior; we show that a scalar reliability cannot move mass to a neighboring class and so cannot correct a directional error. An unrestricted $K\times K$ matrix or Dirichlet calibration can, but spends $K(K-1)$ or $K(K+1)$ parameters on the few labels available. We propose CORDIAL (\underline{\textbf{C}}alibrating \underline{\textbf{ordi}}n\underline{\textbf{al}} outputs), a structured semantic channel whose five parameters, temperature, offset, gain, concentration, and strength, match the geometry of ordinal errors, in a mixture form that keeps the shape of the reading and a unimodal location form. With five parameters the posterior can be averaged over, which hedges against the estimation error that dominates with a few dozen labels, and the channels are stacked with logit-space calibrators.

Our contributions are: (C1) the Bayesian structured channel in two forms, a proof that it never reverses the ordinal order of readings, and its stacking with logit-space calibrators; (C2) a parameter-counting crossover, $n^\star=(d_{\rm u}-d_{\rm s})/(2\delta)$ for a measurable gap $\delta$, that predicts the scale at which unrestricted calibrators take over; and (C3) three uses of the five-parameter posterior, a prior learned across tasks, few-label fusion of several frozen LLMs, and a one-parameter per-user shift, with evidence on Amazon Reviews 2023 and CMU-MOSEI transcripts with four readers.

\noindent\textbf{Related work.}
Temperature, vector, and Dirichlet scaling calibrate class probabilities from held-out labels~\cite{guo2017calibration,kull2019dirichlet} and are the usual fix for miscalibrated LLM confidences~\cite{tian2023just}, and contextual and batch calibration~\cite{zhao2021calibrate,zhou2024batch} remove an LLM's label bias without labels; all ignore the order of the classes. Ordinal-aware losses calibrate classifiers during training~\cite{kim2024ordinal}, and LLM raters on ordinal scales drift toward the middle~\cite{zhang2026auditing} and depend on the order of options and demonstrations~\cite{wang2026positional}; we correct a frozen reader after the fact. Confusion-matrix models of noisy observers go back to Dawid and Skene~\cite{dawid1979maximum}, with ordinal crowd labels aggregated by minimax conditional entropy~\cite{zhou2014ordinal}, and human and model probabilities fused through calibrated confusion matrices~\cite{kerrigan2021combining}. Ordinal regression~\cite{mccullagh1980regression} maps features directly to the scale; supra-Bayesian pooling treats an expert's statement as data with its own likelihood~\cite{genest1986combining}, which is the view taken here for an LLM.

\section{CORDIAL: Structured Channels}
\label{sec:method}

\subsection{LLM outputs as noisy ordinal measurements}
A LLM reads a text $x$ and returns option logits $\ell(x)$ over $K$ ordered classes $y\in\{0,\ldots,K-1\}$, taken from the next-token scores of the answer options. The quantity of interest is the true ordinal label $z$. We treat the reading $q(y\mid x)=\softmax(\ell/\tau)_y$ as a noisy measurement of $z$ and calibrate it through a row-stochastic channel $T$,
\begin{equation}
\hat p(z\mid x)=\sum_{y}T(z\mid y)\,q(y\mid x).
\label{eq:composition}
\end{equation}
The LLM stays frozen; estimating $T$ from a few labeled texts is the calibration problem. The temperature $\tau$ belongs to the channel because frozen readers are saturated: on our benchmarks the largest probability of $\softmax(\ell)$ averages 0.92--0.96.

\subsection{Why a scalar reliability is not enough}
The usual channel trusts the reading with probability $\beta$ and otherwise falls back on the label prior $\pi$, $T_\beta(z\mid y)=\beta\,\ind\{z=y\}+(1-\beta)\pi(z)$, so that $\hat p_\beta=\beta q+(1-\beta)\pi$.

\noindent\textbf{Remark 1 (scalar rigidity).} Since $\hat p_\beta-\pi=\beta(q-\pi)$, a scalar channel only rescales the reading's excess over the prior and cannot move it to a neighboring class. If a saturated reading $q=e_y$ has label $y+1$, every scalar channel assigns that label $(1-\beta)\pi(y+1)\le\pi(y+1)$, no more than ignoring the reader.

\subsection{An affine channel with interpretable parameters}
Ordinal readers err in structured ways: by a consistent offset, by compressing or stretching the scale, and by a consistent spread. With a Gaussian profile on the scale,
\begin{equation}
G(z\mid c)=\frac{\exp[-\kappa\,(z-c)^2]}{\sum_{v}\exp[-\kappa\,(v-c)^2]},
\label{eq:kernel}
\end{equation}
we use two forms of one affine channel,
\begin{align}
\hat p_{\rm mix}&=\rho\sum_y q(y\mid x)\,G(z\mid s\,y+d)+(1-\rho)\,\pi(z),\label{eq:mix}\\
\hat p_{\rm loc}&=\rho\,G\big(z\mid s\,m(x)+d\big)+(1-\rho)\,\pi(z),\label{eq:loc}
\end{align}
where $m(x)=\sum_y y\,q(y\mid x)$ is the reading's mean. The parameters are $\phi=(\tau,d,s,\kappa,\rho)$: temperature, offset $d$, gain $s>0$ ($s>1$ stretches a compressed reading, $s<1$ contracts an exaggerated one), concentration $\kappa>0$, and strength $\rho$ of the reading against the smoothed label histogram $\pi$. The mixture form~\eqref{eq:mix} keeps the shape of the reading, and $d=0$, $s=1$, $\kappa\to\infty$ recovers the scalar channel; the location form~\eqref{eq:loc} is unimodal, which suits labels that are rounded averages. A vector-scaled variant replaces $\ell/\tau$ by $e^{\epsilon}\odot\ell/\tau+b$ with $2K$ more parameters shrunk toward zero.

\noindent\textbf{Proposition 1 (ordinal coherence).}
\textit{Write $q_1\preceq q_2$ (first-order stochastic dominance) if $\sum_{z\le t}q_1(z)\ge\sum_{z\le t}q_2(z)$ for all $t$. For any $\kappa>0$, $s>0$, $d$, $\rho\in[0,1]$, and $\pi$, both forms satisfy $q_1\preceq q_2\Rightarrow\hat p_1\preceq\hat p_2$. The same holds for the posterior predictive provided that the tempered readings remain ordered for every temperature in the posterior support.} Saturated readings of ordered classes meet this condition at every temperature.
Calibration can change a reading's confidence, bias, and spread but cannot turn a higher reading into a lower prediction; unrestricted calibrators can (Section~3.1).

\noindent\textit{Proof.} (a) For $c_1<c_2$, $G(z\mid c_2)/G(z\mid c_1)\propto e^{2\kappa(c_2-c_1)z}$ increases in $z$, so $G(\cdot\mid c)$ increases in likelihood-ratio, hence stochastic, order in $c$~\cite{shaked2007stochastic}. (b) Mixture form: $c(y)=sy+d$ increases in $y$, so $\sum_{z\le t}G(z\mid c(y))$ decreases in $y$, and its average is larger under $q_1$ than under $q_2$. Location form: $m(q)=\sum_z z\,q(z)$ increases under $\preceq$, so $s\,m(q_1)+d\le s\,m(q_2)+d$ and (a) applies. (c) Mixing with the same $\pi$ or averaging over the same posterior keeps the order. $\square$

\subsection{Bayesian estimation from few labels}
With five parameters, unlike with an unrestricted matrix, the posterior can be averaged over cheaply. We place independent Gaussian priors on $(\log\tau,d,\log s,\log\kappa,\operatorname{logit}\rho)$ centered at a near-identity channel ($\tau=2$, $d=0$, $s=1$, $\kappa=4$, $\rho=0.95$) with standard deviations 1, 1, 0.5, 1.5, and 2, and $\epsilon\sim\mathcal N(0,0.5^2)$, $b\sim\mathcal N(0,1)$. We predict with the posterior mean of $\hat p$ over 300 importance-reweighted draws from the Laplace approximation at the mode (covariance inflated by $1.5^2$). The two forms on the two readings give four channels.

\subsection{Stacking with logit-space calibrators}
The channels are combined with three calibrators that act on the logits: temperature scaling, vector scaling~\cite{guo2017calibration}, and proportional-odds regression~\cite{mccullagh1980regression} on the centered logits with an $L_2$ penalty of $100/n$,
\begin{equation}
\hat p(z\mid x)=\sum_{m=1}^{7}\omega_m\,\hat p_m(z\mid x),
\label{eq:stack}
\end{equation}
with weights $\omega$ that maximize the out-of-fold log-likelihood on four folds of the $n$ labels under a Dirichlet(1) prior, found by EM; with fewer than eight labels the weights are equal. The same stack without channels (the \emph{logit stack}), with Bayesian scalar channels, or with full $K\times K$ channels shrunk toward the scalar one by cross-validated KL penalties isolates what the structure adds. Channel forms and priors were chosen on validation texts from the calibration pool.

\subsection{When does structure pay?}
Let $\delta$ be the approximation gap of the structured family, the expected log loss of its best member minus that of the best unrestricted calibrator, and $d_{\rm s}=5$, $d_{\rm u}$ the parameter counts. Under the regularity conditions for maximum-likelihood estimation of possibly misspecified models~\cite{white1982misspecified}, each estimator's expected excess log loss is its gap plus about $d/(2n)$, so structure is expected to win when
\begin{equation}
n<n^\star=\frac{d_{\rm u}-d_{\rm s}}{2\delta}.
\label{eq:crossover}
\end{equation}
Higher-order terms and shrinkage move the actual crossover, so~\eqref{eq:crossover} predicts its scale rather than bounding it; Section~3 checks this with a known $\delta$.

\subsection{What a five-parameter posterior enables}
\noindent\textbf{A learned prior.}
The parameters mean the same thing on every task, so their prior can be learned from other tasks by empirical Bayes. We fit each channel on the full calibration pools of the other datasets, with every reader, express $d$ and $\log\kappa$ in units of the scale length $K-1$, and use the mean of these fits with standard deviation $\min(\sigma_0,\sqrt{v+\sigma_0^2/4})$, where $v$ is their variance and $\sigma_0$ the default.

\noindent\textbf{Several readers.}
$R$ frozen LLMs are $R$ sensors of one label. Each reader gets its own stack~\eqref{eq:stack}, and the $R$ stacked predictions are combined by a second out-of-fold stack together with a multi-reader location channel $G(z\mid s\sum_r w_r m_r(x)+d)$, with a temperature per reader and weights $w$ on the simplex. This channel has $2R+3$ parameters and is estimated in the same Bayesian way as the single-reader channels.

\noindent\textbf{Personalization.}
For repeated texts from one author $u$, the affine channel with $s=\rho=1$, offset $\Delta_u$, and $\kappa\to\infty$ shifts the population prediction by $\Delta_u$ classes (clipped at the ends); $\Delta_u$ maximizes the likelihood of $u$'s earlier labeled texts minus $\Delta_u^2/(2\sigma^2)$, with $\sigma$ cross-validated over users. By Remark~1, a per-user reliability could only rescale $u$'s evidence.

\begin{figure*}[t]
\centering
\includegraphics[width=\textwidth]{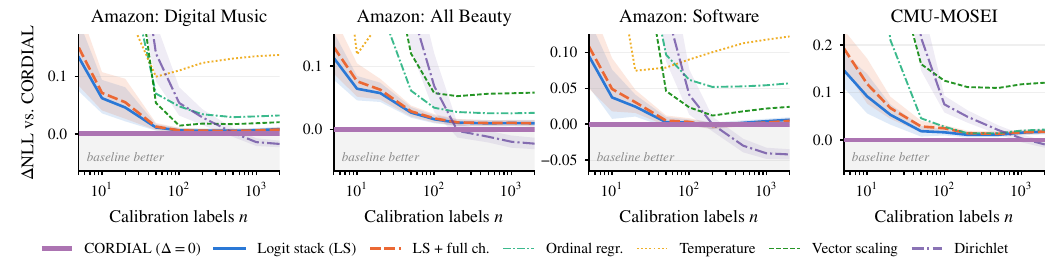}
\caption{Test NLL of each calibrator minus that of CORDIAL (above zero: CORDIAL better) against the number of calibration labels $n$, 7B reader, means over 20 label subsets; bands: paired 95\% bootstrap intervals for the three strongest baselines. The logit stack (LS) combines temperature scaling, vector scaling, and ordinal regression. Curves above a panel's range are cut.}
\label{fig:sweep}
\end{figure*}

\section{Experiments}
\label{sec:experiments}

\noindent\textbf{Benchmarks.}
Amazon Reviews 2023~\cite{hou2024amazon} provides review text, a 1--5 star rating ($K=5$), and a reviewer identifier; we use three product domains (Digital Music, All Beauty, Software), each with 2,000 calibration and 4,000 report reviews from single-review users plus 400 repeat users with at least eight reviews (up to 16 each, in time order). CMU-MOSEI~\cite{zadeh2018mosei} provides transcribed video segments with an averaged sentiment in $[-3,3]$, rounded to seven levels ($K=7$); its train and validation splits (18,135 segments) form the calibration pool and its test split (4,653 segments from 675 videos) the report pool.

\noindent\textbf{Readers.}
Frozen Qwen2.5-Instruct 3B, 7B (main), and 14B~\cite{yang2024qwen} and Llama-3.1-8B-Instruct~\cite{dubey2024llama} are prompted with the text and the $K$ options. The 7B reader's mode is right for 75\%, 66\%, and 68\% of Amazon reviews and 32\% of MOSEI segments, and its errors are directional (Fig.~\ref{fig:reliability}).

\noindent\textbf{Methods and protocol.}
All calibrators use the same $n\in\{5,10,\ldots,2000\}$ calibration labels (20 draws each) and are scored on the report pool by NLL and the ranked probability score (RPS), with probabilities floored at $10^{-4}$. Baselines are temperature and vector scaling~\cite{guo2017calibration}, proportional-odds regression on the centered logits~\cite{mccullagh1980regression} with a cross-validated $L_2$ penalty, Dirichlet calibration~\cite{kull2019dirichlet} with cross-validated off-diagonal and intercept penalties, and the three stacks of Section~2.5. Label-free batch calibration~\cite{zhou2024batch} does not lower the raw NLL. Paired 95\% intervals use 10,000 bootstrap draws over report items and label subsets; a difference is resolved when its interval excludes zero. Fitting ours takes 7--30~s on one CPU core and predicting 0.2~ms per text.

\subsection{Few labels: structure beats size}
\begin{figure}[t]
\centering
\includegraphics[width=\columnwidth]{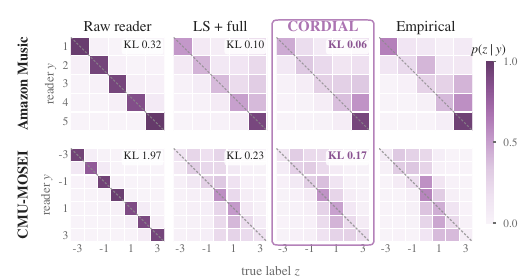}
\caption{Label distributions given the 7B reader's predicted class $y$ (rows) over the true label $z$ (columns; dashed: $z=y$), 20 labels, mean over 20 subsets. Empirical: observed label frequencies. The raw reading is nearly a point mass at $z=y$; the labels shift up on Amazon and toward neutral on MOSEI.}
\label{fig:reliability}
\end{figure}

With the 7B reader, ours has the lowest NLL among nine calibrators, the eight in Table~\ref{tab:sweep} and the fixed-penalty ordinal regression used inside the stack, at every budget from 5 to 100 labels on all four benchmarks (Fig.~\ref{fig:sweep}). With 20 labels it matches the logit stack with 28--54 labels and Dirichlet calibration with 79--132. Against the logit stack it gains 0.046, 0.057, 0.024, and 0.053 nats at $n=20$ for Music, Beauty, Software, and MOSEI and 0.002--0.016 at $n=100$, with every gain resolved except on Software beyond 20 labels; the same stack with full or scalar channels changes the logit stack's NLL at $n=20$ by $-0.015$ to $+0.014$ and trails ours by resolved margins of 0.031--0.067, so the gain comes from the structure rather than from more members. Posterior averaging adds up to 0.134 nats over the posterior mode with ten labels or fewer and nothing from 100 on. The RPS agrees, favoring ours over the logit stack by 0.008--0.017 at $n=20$, and accuracy matches the logit stack. Grouped by the reader's stated class (Fig.~\ref{fig:reliability}), ours is closest to the labels on all four benchmarks at $n=20$: its class-weighted KL is 0.06--0.17, against 0.10--0.22 for the logit stack, 0.31--0.55 for Dirichlet calibration, and 0.32--1.97 for the raw reader. The labels are ordered in $y$ on every benchmark, and so, by Proposition~1, are the temperature-reading channels; fitted on 20 labels, vector scaling and Dirichlet calibration map saturated readings of some adjacent classes to predictions reversed by more than 0.05 in cumulative probability in 45--80\% and 45--70\% of fits, and ours in at most 10\%.

\begin{table}[t]
\centering
\caption{Test NLL at $n=20$ and $n=200$ calibration labels, 7B reader (means over 20 subsets; bold: best). CORDIAL adds the structured channels to the logit stack (LS).}
\label{tab:sweep}
\small
\setlength{\tabcolsep}{1.5pt}
\begin{tabular}{@{}l rr rr rr rr@{}}
\toprule
& \multicolumn{2}{c}{Music} & \multicolumn{2}{c}{Beauty} & \multicolumn{2}{c}{Software} & \multicolumn{2}{c}{MOSEI} \\
\cmidrule(lr){2-3}\cmidrule(lr){4-5}\cmidrule(lr){6-7}\cmidrule(l){8-9}
Method & 20 & 200 & 20 & 200 & 20 & 200 & 20 & 200 \\
\midrule
Raw LLM & 1.339 & 1.339 & 2.052 & 2.052 & 2.093 & 2.093 & 5.806 & 5.806 \\
Temperature & 0.693 & 0.544 & 0.766 & 0.743 & 0.792 & 0.763 & 1.682 & 1.662 \\
Vector scaling & 0.907 & 0.438 & 0.953 & 0.593 & 1.023 & 0.675 & 1.729 & 1.283 \\
Ordinal regr. & 0.787 & 0.454 & 0.789 & 0.568 & 1.028 & 0.715 & 1.540 & 1.186 \\
Dirichlet & 0.982 & 0.449 & 1.181 & \textbf{0.539} & 1.308 & 0.663 & 2.020 & 1.221 \\
Logit stack (LS) & 0.519 & 0.426 & 0.654 & 0.551 & 0.742 & 0.661 & 1.383 & 1.183 \\
LS + scalar ch. & 0.504 & 0.429 & 0.662 & 0.554 & 0.749 & 0.664 & 1.395 & 1.187 \\
LS + full ch. & 0.528 & 0.426 & 0.659 & 0.552 & 0.748 & \textbf{0.660} & 1.397 & 1.186 \\
CORDIAL & \textbf{0.473} & \textbf{0.420} & \textbf{0.596} & 0.541 & \textbf{0.718} & 0.663 & \textbf{1.330} & \textbf{1.172} \\
\bottomrule
\end{tabular}
\end{table}

\noindent\textbf{The crossover.}
Dirichlet calibration, the strongest unrestricted calibrator, crosses zero in Fig.~\ref{fig:sweep} and first leads with a resolved interval at 1000, 500, 500, and 2000 labels and at $n=2000$ by 0.018, 0.022, 0.042, and 0.009 nats. Using the large-$n$ gaps as plug-in estimates of $\delta$, Eq.~\eqref{eq:crossover} with $d_{\rm u}=K(K+1)$ gives 
$n^\star=700$, 565, 297, and 2746, all within a factor of two 
of the observed crossovers and preserving their ordering. With labels from a known channel outside the affine family ($\delta=0.001$--$0.09$), maximum-likelihood excess losses follow $d/(2n)$ with $d\approx4$ for the four-parameter channel and $1.1$--$1.3\,K(K-1)$ for the full matrix, overtaking at 1.2--2.7$\,n^\star$ without shrinkage and 0.2--1.3$\,n^\star$ with it.

\noindent\textbf{Reader size and family.}
With the 3B, 14B, and Llama readers, CORDIAL is best in 56 of 60 settings with 5 to 100 labels, trails by at most 0.006 in the other four, and leads the logit stack at $n=20$ by resolved margins of 0.012--0.086.

\noindent\textbf{In-context examples.}
Used as in-context demonstrations for the 7B reader, the same 20 to 100 labels give accuracy within $-0.06$ to $+0.02$ of ours but a 2.3 to 3.3 times higher NLL.

\subsection{What the posterior buys}
\noindent\textbf{A learned prior.}
With the prior learned from the other datasets, ours gains on average 0.017, 0.016, 0.011, 0.006, and 0.003 nats with 5, 10, 20, 50, and 100 labels over the twelve Amazon reader--domain pairs; the gain is resolved for 9 or 10 pairs at each budget, and no pair shows a resolved loss. On MOSEI, with a prior from Amazon alone, it costs 0.009--0.011 at $n\le10$ and gains 0.013--0.015 at 20--50.

\noindent\textbf{Several readers.}
The four-reader fusion, designed on validation texts, beats the 7B reader alone in all 28 settings from 5 to 500 labels, by 0.008--0.090 nats, the reader chosen by cross-validation in all 28, and the same fusion without channels in 27, by 0.011--0.043 at $n=20$; Dirichlet calibration of the concatenated readings trails by 0.55--1.5 nats there. On MOSEI with 10--20 labels Llama alone is 0.018--0.030 better; the multi-reader channel adds at most 0.006.

\noindent\textbf{Across users.}
A per-user shift channel on the population fit (1,000 labels) gains, after eleven or more earlier texts of the user, 0.100, 0.046, 0.044, and 0.048 nats on Music, Beauty, Software, and MOSEI, all resolved. Over all texts it beats a per-user reliability by 0.013--0.041, a per-user temperature by 0.004--0.016, and a random-intercept ordinal regression by 0.021--0.029 on Amazon, tying it on MOSEI.

\section{Conclusion}
An LLM’s ordinal output can be viewed as a noisy sensor with systematic offset, gain, and concentration errors. Our five-parameter Bayesian channel calibrates these distortions from few labels and naturally extends to learned priors, multiple readers, and user-specific adaptation.

\clearpage
\bibliographystyle{IEEEbib}
\bibliography{references}

\section{Compliance with Ethical Standards}
This study reuses two public datasets, Amazon Reviews 2023 and CMU-MOSEI transcripts, with the user and speaker identifiers they provide. It involves no new participant recruitment or intervention.
\end{document}